\documentclass[11pt,a4paper]{article}
\usepackage{times,latexsym}
\usepackage{url}
\usepackage[T1]{fontenc}

\usepackage[acceptedWithA]{tacl2021v1}

\usepackage{xspace,mfirstuc,tabulary}

\newif\iftaclinstructions
\taclinstructionsfalse 
\iftaclinstructions
\renewcommand{\confidential}{}
\renewcommand{\anonsubtext}{(No author info supplied here, for consistency with
TACL-submission anonymization requirements)}
\newcommand{\instr}
\fi

\iftaclpubformat 

\else

\fi

\usepackage{enumitem}
\setlist[enumerate,1]{leftmargin=1.2em,labelindent=0em,itemsep=0pt,labelsep*=0.5em}

\title{Interactive Machine Teaching by Labeling Rules and Instances}

\author{
  Giannis Karamanolakis\Thanks{Work done at Columbia prior to joining Amazon.} 
  \\
  Amazon AGI
  \\
  New York, NY 10001, USA
  \\
  \texttt{{\normalsize karamai@amazon.com}}
  \And
  Daniel Hsu 
  \\
  Columbia University
  \\
  New York, NY 10027, USA
  \\
  \texttt{{\normalsize djhsu@cs.columbia.edu}}
    \And
  Luis Gravaano 
  \\
  Columbia University
  \\
  New York, NY 10027, USA
  \\
  \texttt{{\normalsize gravano@cs.columbia.edu}}
}

\date{}

\usepackage{wrapfig}
\usepackage{booktabs,array,arydshln} 
\usepackage{graphicx}
\usepackage{subcaption}  
\usepackage{comment}
\usepackage[utf8]{inputenc}
\usepackage{amsmath}
\usepackage{amsfonts}
\usepackage{mathtools}
\usepackage{mathptmx}
\usepackage{bbm}
\usepackage{algorithm}
\usepackage{algpseudocode}

\newcommand{\iwsysname}{{\mbox{INTERVAL}}\xspace}

\definecolor{forestgreen}{rgb}{0.13, 0.55, 0.13}
\newcommand{\segment}{s}

\newcommand{\softlabel}{\textbf{p}}
\newcommand{\isoftlabel}{p}

\newcommand{\iteachersoftlabel}{q}
\newcommand{\hardlabel}{y}
\newcommand{\numclasses}{K} 

\newcommand{\featurespace}{\mathcal{X}}
\newcommand{\labelset}{\mathcal{Y}}
\newcommand{\labelspace}{\labelset}

\newcommand{\abstain}{\bot}

\usepackage{pifont}
\begin{document}
\maketitle
\begin{abstract}
    Weakly supervised learning aims to reduce the cost of labeling data by using expert-designed labeling rules. 
    However, existing methods require experts to design effective rules in a single shot, which is difficult in the absence of proper guidance and tooling.
    Therefore, it is still an open question whether experts should spend their limited time writing rules or instead providing instance labels via active learning. 
    In this paper, we investigate how to exploit an expert's limited time to create effective supervision.
    First, to develop practical guidelines for rule creation, we conduct an exploratory analysis of diverse collections of existing expert-designed rules and find that rule precision is more important than coverage across datasets. 
    Second, we compare rule creation to individual instance labeling via active learning and demonstrate the importance of both across 6 datasets.
    Third, we propose an interactive learning framework, \iwsysname, that achieves efficiency by automatically extracting candidate rules based on rich patterns (e.g., by prompting a language model), and effectiveness by soliciting expert feedback on both candidate rules and individual instances. 
    Across 6 datasets, \iwsysname outperforms state-of-the-art weakly supervised approaches by 7\% in F1. Furthermore, it requires as few as 10 queries for expert feedback to reach F1 values that existing active learning methods cannot match even with 100 queries. 
\end{abstract}

\section{Introduction}
\label{s:motivation-interactive-weak-supervision}
Supervised machine learning models for text classification require large, hand-labeled training datasets, which are both expensive and time-consuming to obtain.
Most efforts to reduce the reliance on large training datasets support just a single type of expert supervision, namely to label individual instances one at a time~\cite{seeger2006taxonomy,clark2018semi,ruder2018strong,berthelot2019mixmatch,peters2018deep,devlin2019bert,zhang2021survey,zhang-etal-2022-survey}.

To reduce the data labeling bottleneck, weakly supervised learning (WSL)~\cite{zhang2022survey} focuses on labeling rules that automatically generate weak labels for unlabeled instances.
WSL works in two separate steps: (i) experts provide labeling rules; and (ii) labeling rules are used to train a machine learning model.
Most work focuses on solving the second step and learn with noisy rules~\cite{ratner2016data,ratner2017snorkel,karamanolakis2019cotraining,bach2019snorkel,awasthi2019learning}.
In practice, however, experts find it difficult to define sufficiently many rules in one shot~\cite{varma2018snuba}.
Considerable time and creativity are required for inspecting unlabeled instances and creating rules that add predictive value by effectively covering a substantial number of instances.
Therefore, it is an open question whether experts should spend their limited time writing rules or instead providing instance labels, notably via active learning~\citep{settles2009active}. 

In this paper, we investigate how to efficiently exploit an expert's limited time for machine teaching.
Our main idea is to \emph{automatically} extract labeling rules with high coverage of unlabeled data, and then rely on domain expertise to validate the candidate rules.
In contrast to active learning methods where the machine queries the expert for labels of individual examples~\cite{zhang-etal-2022-survey}, providing feedback for each rule leads to multiple data labels, which we show here can boost classification performance faster. 

Supporting rich forms of interaction is challenging, especially when the teaching budget is limited.
First, given a restricted number of rules that can be created or validated by an expert, it is not clear what properties these rules should have to train an accurate model. 
For example, should one prioritize rules that cover many examples but with relatively low precision, or rules that have high precision but lower coverage?
Moreover, existing algorithms for rule extraction require substantial labeled data, and it is unclear how to extract and rank candidate rules when we are given just limited labeled data and perhaps a few expert-validated rules. 
In general, there are few guidelines in the literature for creating effective rules for efficient machine teaching.
Additionally, the option to ask for feedback on both rules and instances requires balancing the costs and potential benefits of each type of feedback when there is a shared budget of expert interaction. 

Our work addresses these open questions via the following contributions: 

\paragraph{Characterization of prevalent patterns in offline machine teaching.} We analyze six datasets with expert-defined rules and evaluate multiple weak supervision methods under simulated low-resource settings.
Specifically, we unify several weak supervision methods using a Teacher-Student abstraction, where a subset of the rules are considered in the teacher model for training a student model. 
 By evaluating more than 1,000 Teacher-Student configurations per dataset, we associate Teacher properties with the Student's performance and, even though rules are dataset-specific, we find two prevalent patterns across datasets and methods that could inform guidelines for rule creation. First, we show that a higher-F1 Teacher does not necessarily lead to a higher-F1 Student. Second, we show the Teacher's precision is more important than coverage for training an accurate Student.

\begin{figure}
    \centering
    \includegraphics[width=\columnwidth]{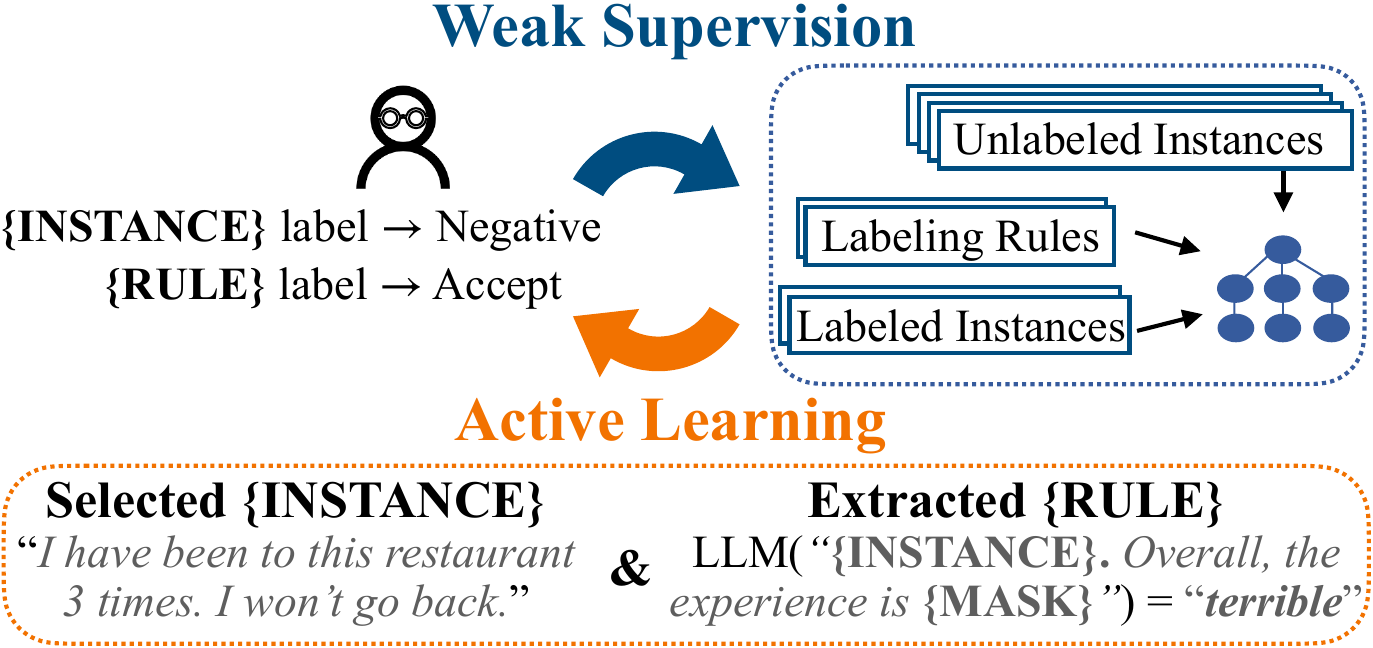}
    \caption{Our INTERVAL framework supports interaction on both instances and automatically-extracted rules (e.g., by prompting a large language model) for weakly supervised learning.}
    \label{fig:interval_intro_figure}
\end{figure}

\paragraph{Automatic rule extraction via prompting.} 
We propose a method that extracts rules with rich predicates, expressed as conjunctions of $n$-grams, syntactic features, and prompt-based features. 
By prompting a pre-trained model (see Figure~\ref{fig:interval_intro_figure}), our method extracts 
high-level features that might not explicitly appear in the text (e.g., ``terrible'' customer experience) and thus can discover common patterns across instances with no $n$-gram overlap.
As we will show, by extracting both surface-level and higher-level features, our rule family achieves higher precision and coverage than $n$-gram rules.
Our design focuses on rules that could be easily validated by a human and are highly effective.

\paragraph{Interactive machine teaching.} We present a human-in-the-loop machine teaching framework called \iwsysname,\footnote{\iwsysname: INTEractive Rule discoVery for weAkly supervised Learning.} which queries for expert feedback on both instances and rules, and uses all the available resources to train a classifier. We quantify the trade-off between labeling rules vs. instances and show that our framework is more efficient than existing WSL and active learning approaches even when starting with no expert-written rules. 
Our analysis demonstrates that feedback on both rules and instances is more effective than feedback on instances only (as in Active Learning) even when labeling rules are more expensive than labeling instances by up to 9 times.

The rest of this paper is organized as follows. Section~\ref{s:iws-related-work-interactive-machine-teaching} reviews related work on interactive machine teaching and defines our problem of focus. Section~\ref{s:iws-method} presents our interactive machine teaching framework,\footnote{Our implementation is publicly available at \url{https://github.com/gkaramanolakis/interval}.} which queries for feedback on labeling rules and instance. Sections~\ref{s:iws-experimental-settings} and~\ref{s:iws-experimental-results} evaluate our interactive method via experiments on six text classification datasets. 
Finally, Sections~\ref{s:interval_discussion} and ~\ref{iws:conclusions} conclude and suggest future work.

\section{Problem Definition and Related Work}
\label{s:iws-related-work-interactive-machine-teaching}
We now define our problem of focus (Section~\ref{s:iws-problem-definition}); we also discuss related work on non-interactive weak supervision and interactive learning with instance- and feature-level feedback  (Section~\ref{s:iws-prior-work})

\subsection{Problem Definition}
\label{s:iws-problem-definition}
Let $\mathcal{X}$ denote the feature space and $\labelspace= \{1, \dots , \numclasses\}$ denote the label space for a $\numclasses$-class classification task. 
We consider a set of manually-labeled examples $D_L = \{(\segment_l, \hardlabel_l)\}$, where $\segment_l \in \featurespace$ and $\hardlabel_l \in \labelspace$, and a set of unlabeled examples $D_U = \{\segment_i\}$.
We also consider a set of pre-defined expert-provided labeling rules $R = \{r^j\}$.
A rule $r^j: \mathcal{X} \rightarrow \labelspace \cup \{\abstain\}$ maps an example $\segment_i$ into a label $z_i^j \in \labelspace \cup \{\abstain\}$.
Predicting $z_i^j=\abstain$ indicates that $r^j$ does not cover $\segment_i$.
We are primarily interested in the scenario where the size of $D_L$ is small in comparison to that of $D_U$, and where $R$ contains just a few or no expert-provided rules, which is often the case for new tasks.
Additionally, we assume that we have a budget of $T$ ``cost'' units (e.g., time) for querying a subject matter expert for feedback on either an instance $\segment_i \in D_U$ (at a cost of $T_I$) or an automatically extracted rule $r^j$ (at a cost of $T_R$), as we discuss in Section~\ref{s:iws-prior-work}.

Our goal is to leverage $D_L$, $D_U$, and $R$, and interact with the expert within the specified budget $T$ to train a classifier that, given an unseen test instance $\segment' \in \featurespace$, predicts a label $\hardlabel'\in \labelspace$.

\subsection{Prior Work}
\label{s:iws-prior-work}
\paragraph{Non-interactive approaches.}
\label{s:iws-related-work-non-interactive}
Non-interactive weak supervision approaches do not involve a human in the loop (i.e., $T=0$ for our problem definition).
Supervised learning methods consider just $D_L$, semi-supervised learning methods consider $D_L$ and $D_U$~\cite{nigam2000analyzing,lee2013pseudo,gera2022zero},
and WSL methods consider $D_L$, $D_U$, and $R$~\cite{ratner2017snorkel,bach2019snorkel,badene2019data,fu2020fast,awasthi2019learning,karamanolakis2021astra}. 
WSL uses rules in $R$ (e.g., keyword-based patterns, regular expressions, heuristic labeling functions) to automatically generate weak training labels for unlabeled instances in $D_U$.
As rules can be noisy, can have limited coverage, and different rules may generate conflicting labels for the same instance, WSL techniques estimate rule weights for noise-aware training~\cite{zhang2022survey}.  Our method also employs WSL, can work with any rule-weighting technique and further discovers new rules to expand the coverage of $R$. 

Our method is also related to zero-shot and few-shot prompting methods, which use a template to modify the input $\segment_i$ into a cloze-style or entailment question and leverage a pre-trained model to ``answer'' the question~\cite{schick2021exploiting,yin2019benchmarking,liu2023pre}.
By directly using the outputs of the pre-trained model for classification, prompt-based techniques are sensitive to the selection of prompting templates~\cite{gao2021making,ye2023comprehensive}, labeled examples~\cite{zhao2021calibrate,perez2021true}, and hyperparameters~\cite{tam2021improving}. Even prompting powerful models such as ChatGPT, the successor of InstructGPT~\citep{ouyangtraining}, requires work to reach the performance of supervised (fine-tuned) models on text benchmarks~\citep{bang2023multitask}.
Our work explores prompting for rule creation during training instead of direct inference.
Specifically, we use the pre-trained model's output to construct labeling rules, which we assume are only weakly indicative of the true labels. 
Through our approach prompting is required just for training and can work with any model for inference, thus enabling applications where deploying large language models might not be possible. 

Our work is also related to rule extraction methods, which consider rules of various types such as keywords, named entities and numeric expressions~\cite{yangarber2000unsupervised}, synthetic relations~\cite{snow2004learning}, part-of-speech tags and hypernyms~\cite{califf2003bottom}, regular expression patterns~\cite{augenstein2016stance}, sequential patterns~\cite{srikant1996mining, jindal2008opinion}, and more recently, features extracted by prompting pre-trained models~\cite{zhang2022prompt}.
Our method considers a rich family of rules based on $n$-grams, linguistic features (e.g., part of speech tags and named entities), and prompt-based features and focuses on efficient interaction by soliciting feedback on both candidate rules and instances.

\paragraph{Interactive learning with instance feedback.}
\label{s:iws-related-work-active-learning}
One type of interaction that has been studied extensively in the literature is active learning, in which the machine queries the expert for just a small number of labels for examples that are chosen adaptively from abundant unlabeled data~\cite{lewis1994sequential,cohn1996active,roy2001toward,dasgupta2007general,dasgupta2008hierarchical,settles2009active,beygelzimer2010agnostic,houlsby2011bayesian,zhang2015active,shen2017deep,kirsch2019batchbald,ash2019deep,brantley2020active,yuan2020cold,dor2020active,margatina2021active,zhang-etal-2022-survey}. 
Nearly all previous active learning methods solicit the expert's judgment to just label instances. 
In other words, they do not support feedback on labeling rules (i.e., $T_R=\infty$) and query for feedback on $\lfloor \frac{T}{T_I} \rfloor$ instance labels. 
Creating a sufficiently large training set would require separate feedback on many individual instances.
On the other hand, validating a candidate \emph{rule} leads to weak labels for \emph{many} examples at a time (i.e., for all the examples covered by the rule) and, as a result, a large weakly-labeled dataset can be created with a relatively small number of rules. 

\paragraph{Interactive learning with rule feedback.}
\label{s:iws-related-work-iws}
Our work is related to previous interactive methods that support expert queries on automatically
generated rules from the $n$-gram family~\cite{druck2008learning,melville2009sentiment,settles2011closing,jagarlamudi2012incorporating,poulis2017learning,dasgupta2018learning,boecking2020interactive,kartchner2022rule}. 
These methods extract simple $n$-gram based rules, which as we will show (e.g., in Figure~\ref{fig:autorule_scatterplots}) have limited effectiveness and different characteristics than expert-provided rules in $R$.
As two exceptions,~\citet{sen2019heidl} extracts rules based on linguistic expressions via syntactic parsing and ~\citet{zhang2022prompt} considers rules based on the output of pre-trained language models prompted with task-specific templates; both show that experts can successfully provide feedback on rules from the proposed families.
Most of the above methods do not allow instance-labeling queries (i.e., these methods assume that $T_I=\infty$).
In contrast, our method subsumes and generalizes existing work on rule labeling and active learning by querying an expert for both instances and automatically extracted rules from a new rule family with rich predicates.

\section{Interactive Machine Teaching with Instance and Rule Feedback}
\label{s:iws-method}
This section describes our interactive machine teaching framework, which addresses the problem defined in Section~\ref{s:iws-problem-definition}. 
The core question is, how to efficiently solicit expert feedback for machine teaching given a limited budget $T$.
Our main idea is to balance the quality of instance labels with the efficiency of labeling rules under this low-resource setting.
We propose a framework, \iwsysname, that supports efficient interaction by selecting which instances to label manually and by extracting candidate rules that, when accepted, can automatically generate many additional labels.
\iwsysname can be used with several WSL methods and any learning model. 

In the rest of this section, we describe the individual steps followed by \iwsysname on each iteration, namely Teacher-Student co-training (Section~\ref{iws-method-teacher-student-cotraining}), querying for instance feedback (Section~\ref{iws-method-instance-feedback}), candidate rule extraction (Section~\ref{s:iws-candidate-rule-generation}), 
and querying for rule feedback (Section~\ref{iws-method-rule-feedback}), and then we summarize the main ideas of our interactive machine teaching algorithm (Section~\ref{s:iws-interactive-machine-teaching-algorithm}).

\subsection{Teacher-Student Co-Training}
\label{iws-method-teacher-student-cotraining}
In the first step of each iteration, we use $D_L$, $D_U$, and $R$ to train a model.
This has been the main objective in non-interactive WSL.
Our model training employs the Teacher-Student abstraction by~\citet{karamanolakis2021astra} to unify several WSL methods~\cite{dawid1979maximum,ratner2016data,ratner2019training,zhang2022survey}.

The teacher model $q_{\phi}(\cdot)$ considers $D_L$, $D_U$, and $R$, and predicts labels $q_i$ for all examples $\segment_i \in D_U$ except for examples covered by no rules in $R$, which are then not covered by the Teacher either.
The student model $p_\theta(\cdot)$ is the base learning model that is trained using $D_L$, $D_U$, and the teacher model by approximately solving the following optimization problem:
\begin{multline}
\min_\theta\ \mathbb{E}_{\segment_l, \hardlabel_l \in D_L} [-\log\ \isoftlabel_\theta (\hardlabel_l \mid \segment_l)] +\\ \lambda \mathbb{E}_{\segment \in D_U} \mathbb{E}_{\hardlabel \sim \iteachersoftlabel_{\phi^*}(\hardlabel \mid \segment)} [-\log\ \isoftlabel_\theta (\hardlabel \mid \segment)],
\label{eq:iws-student-training}
\end{multline}
where $\lambda \in \mathbb{R}$ is a hyper-parameter controlling the relative weight of the manually labeled data (first term) and the weakly labeled data (second term).

The same Teacher-Student abstraction appears across different WSL approaches~\cite{zhang2022survey}, which differ in the teacher model design. 
For example, in simple majority voting, the Teacher aggregates the predictions of rules in $R$. 
In Snorkel~\cite{ratner2017snorkel}, the Teacher is a probabilistic graphical model that estimates weights for rules in $R$ in an unsupervised way.
In ASTRA~\cite{karamanolakis2021astra}, the Teacher is a rule-attention network that aggregates rule labels with instance-specific weights and is co-trained with the Student. 

In our problem of focus, where the size of $D_L$ is small and $R$ contains just a small number of rules, the student model might have far less than satisfying accuracy for our target task. 
Next, we show how to exploit the interaction budget $T$.

\subsection{Querying for Instance Feedback}
\label{iws-method-instance-feedback}
After having trained the Student, \iwsysname queries the label $\hardlabel_i$ for an instance $\segment_i$ from the unlabeled set $D_U$.
To efficiently interact with an expert, we design a method that chooses which instance to query for feedback based on the Student's probabilities, as some instances might be more ``informative'' for the Student than others. \iwsysname identifies a diverse collection of unlabeled instances for which the Student's predicted probabilities have high entropy as explained next. 

\paragraph{Instance clustering.} 
At the beginning of our algorithm, we construct a hierarchical clustering of the unlabeled instances in $D_U$. 
To achieve this, we implement agglomerative clustering using Ward's linkage method, which focuses on minimizing cluster variances. 
For cluster variances, we calculate the Euclidean distances between instances based on instance embeddings, which are computed via pre-trained BERT~\citep{devlin2019bert}. For implementation details see Section~\ref{s:iws-experimental-settings}.

\paragraph{Instance selection.} 
To choose which instances to query, 
\iwsysname applies the Student $p_\theta(\cdot)$ on each unlabeled instance $\segment_i \in D_U$ to get soft labels $\softlabel_i = (\isoftlabel_i^1, \dots, \isoftlabel_i^K)$, where $\isoftlabel_i^k$ represents the Student's predicted probability for assigning $\segment_i$ into the target class $k \in K$.
 We use $D_{Student}=\{(\segment_i, \softlabel_i)\}_{\segment_i \in D_U}$ to define the dataset that is soft-labeled by the Student.
Then, \iwsysname selects sample instances $\segment_i$ via the cluster-adaptive sampling algorithm of~\citet{dasgupta2008hierarchical}, which exploits the hierarchical structure of the data and evaluates cluster informativeness based on the entropy of the Student's predicted probabilities for $\segment_i$ in $D_S$. 
Specifically, the algorithm chooses instances $\segment_i$ from clusters characterized by low label ``purity'', or equivalently, high entropy based on the Student's probabilities $\softlabel_i$. 
This selection is made under the premise that collecting expert labels for these instances will provide valuable information for the subsequent round of Student training. Once a cluster becomes ``pure,'' then the algorithm shifts its focus to another cluster with the goal to acquire a diverse collection of instances.

\paragraph{Instance labeling.} After selecting an instance $\segment_i$, the system queries the expert's label $\hardlabel_i$ at a cost of $T_I$.
At the end of the iteration, the labeled pair $(\segment_i, \hardlabel_i)$ is added in $D_L$ to train the Teacher and Student at the next iteration.

\subsection{Candidate Rule Extraction}
\label{s:iws-candidate-rule-generation}
In contrast to WSL, where experts manually create rules with significant coverage on $D_U$, we propose to automatically extract candidate rules and hopefully reduce the cost of rule creation.
After getting the label $y_i$ for an instance $\segment_i$, we extract candidate rules $r^j$ that predict the same label $y_i$ for $\segment_i$ and have non-trivial coverage in $D_U$.
We first describe the types of rules and then how to extract them.

\begin{table}[]
    \centering
    \resizebox{1\columnwidth}{!}{
    \begin{tabular}{l|l}
    \toprule
      \textbf{Name}    &  \textbf{Prompt Template}\\
     \midrule
   EXPERIENCE & Overall, the experience is [MASK]. [TEXT].\\
   RECOMMEND   & [TEXT]. Would I recommend it? The answer is [MASK].\\\midrule
   ASKS\_FOR & The following SMS message asks for [MASK]:  [TEXT].\\
   IS\_ABOUT & The following SMS message is about [MASK]:  [TEXT].\\

     \bottomrule
    \end{tabular}}
    \caption{\small Examples of templates used to prompt pre-trained models in Yelp (top) and SMS (bottom) for candidate rule extraction.}
    \label{tab:prompt_template_examples}
\end{table}

\begin{table}[]
    \centering
    \resizebox{\columnwidth}{!}{
    \begin{tabular}{l}
    \toprule
     \textbf{Candidate Rules (predicate $\rightarrow$ label)} \\
     \midrule
    PMT-EXPERIENCE=``terrible'' $\rightarrow$ Negative \\
    PMT-EXPERIENCE=``fantastic''$\rightarrow$ Positive \\
    PMT-RECOMMEND=``certainly'' $\rightarrow$ Positive \\\midrule
    PMT-IS\_ABOUT=``prizes'' $\rightarrow$ Spam \\
    NGRAM=``http'' \textbf{AND}  PMT-ASKS\_FOR=``donations'' $\rightarrow$ Spam \\
    NER=``CARDINAL'' \textbf{AND} PMT-ASKS\_FOR=``information''$\rightarrow$ Spam \\
     \bottomrule
    \end{tabular}}
    \caption{\small Examples of rules extracted by our method for Yelp (top) and SMS (bottom). ``NGRAM=$a$'' means that $a$ appears as an $n$-gram in the text. ``NER=$a$'' means that at least one entity of type $a$ exists in the text. ``PMT-$b$=$a$'' means that $a$ appears among the top-$k$ tokens predicted by the pre-trained model to fill in the [MASK] token for a prompt template $b$.}
    \label{tab:rules_high_level_features_examples}
\end{table}

\paragraph{Rule family.} 
Most work on interactive learning with rule feedback has focused on extracting keyword-based labeling rules. These rules have limited expressiveness compared to expert-written rules, which include class-indicative keywords, regular expression patterns, and auxiliary classifiers (e.g., polarity and subjectivity classifiers for spam classification)~\citep{zhang-etal-2022-survey}.
To improve expressiveness without sacrificing interpretability, our method extracts rules $r^j$ whose predicates $v^j(\segment_i)$ are conjunctions of features that can have three different types: $n$-grams ($v^j(\segment_i)$ is true if a specific $n$-gram appears in $\segment_i$), linguistic features (e.g., part-of-speech tags and named entities), and prompt-based features. 
Specifically to construct prompt-based rules, we prompt pre-trained models for $\segment_i$ using templates from         ``PromptSource''~\cite{bach2022promptsource}.
As an example, consider the sentence from Figure~\ref{fig:interval_intro_figure} ($\segment_i$: ``\textit{I have been to this restaurant 3 times. I won't go back}''). We construct ``prompt-based'' predicates by prompting a pre-trained model to fill in the mask in the following template: ``\textit{<$s_i$>. Overall, the experience is} [MASK]'' and extracting the top $k$ tokens (e.g., ``terrible'').
Table~\ref{tab:prompt_template_examples} shows more examples of prompt templates and Table~\ref{tab:rules_high_level_features_examples} lists examples of rules extracted by our method using such templates (extraction details are discussed later). 
 Our approach extracts common patterns across instances that might not even share any $n$-gram features, such as in tasks with short documents.
 As we will see in Section~\ref{s:experiments-automatic-rule-suggestion}, the rules in our expanded family can be substantially more accurate than the simple $n$-gram rules considered in previous work, and yet they are nearly as interpretable.
Note that, at test time, our method does not require access to the above resources as the student model predicts labels directly based on $\segment_i$.

\paragraph{Rule extraction.} 
We extract rules $r$ from the above family, as long as (i) they cover at least $t_{cov}$ examples in $D_U$ \emph{including} $s_i$ and (ii) they have a precision of at least $t_{prec}$ in $D_L$.
Both $t_{cov}$ and $t_{prec}$ are hyper-parameters.
Given the above coverage and precision constraints, we extract conjunctions of features using the Apriori algorithm~\cite{agrawal1994fast}.
Specifically, we first exhaustively search all rules with a single feature from the above family and keep all rules that satisfy all constraints. (The constraint that all rules have to cover $s_i$ with a label $y_i$ is especially strong and allows efficient search.)
Then, we create rules as conjunctions of two features selected before and pick just the resulting rules that satisfy all of the above constraints. 
Our method considers rules with conjunctions of up to $t_{len}$ features, where $t_{len}$ is another hyper-parameter. 
The set $R_C$ contains all candidate rules that are extracted by our method and satisfy our constraints. 

Automatically identifying a good rule is hard with limited labeled data $D_L$. 
For example, a candidate rule $r^j$ with high coverage on $D_U$ might have low coverage in $D_L$ ($D_L$ might contain just a few labeled examples), and therefore it is hard to estimate the true precision of $r^j$.
Therefore, we rely on expert feedback for selected candidate rules from $R_C$, as discussed next. 

\begin{figure}[t]
    \centering
\begin{algorithm}[H]
\caption{Interactive Machine Teaching}
\label{algo:interactive-machine-teaching}
\begin{algorithmic}[1]
\Statex{\textbf{Input:} Small amount of labeled data $D_L$; task-specific unlabeled data $D_U$; small set of weak rules $R$; budget of $T$ cost units for interaction with a subject matter expert}
\Statex{\textbf{Output:}} Student $\isoftlabel_\theta^*(\cdot)$, Teacher $\iteachersoftlabel_\phi^*(\cdot)$, augmented labeled data $D'_L$, augmented set of weak rules $R'$
    \State{Cluster all data $\segment_i \in D_U$ into hierarchical clusters (agglomerative clustering; Ward's linkage; Euclidean distance of instance embeddings)}
    \State{Initialize $D'_L=D_L$, $R'=R$}
    \State{\textbf{Repeat until the budget $T$ runs out:}}
      \begin{algorithmic}
      \State{{\small 3.1:} Train Teacher  $\iteachersoftlabel_\phi^*(\cdot)$ and Student $\isoftlabel_\theta(\cdot)$ using $D_L'$, $D_U$, $R'$}
      \State{{\small3.2:} Apply $p_\theta(\cdot)$ to $\segment \in D_U$ to obtain soft labels: $D_{Student}=\{(\segment_i, \softlabel_i)\}_{\segment_i \in D_U}$}
      \State{{\small 3.3:} Pick a candidate instance $\segment_i \in D_U$ }
      \State{{\small 3.4:} Query the label $\hardlabel_i$ for $\segment_i$} (cost $=T_I$)
      \State{{\small 3.5:} Extract candidate rules $r^j$ that cover $\segment_i$} 
      \State{{\small 3.6:} Query the labels $z^j$ for $\beta_i$ rules $r^j$ } (cost $=\beta_i \cdot T_R$)
      \State{{\small3.7:} Update $D_L' = D_L' \cup \{(\segment_i, \hardlabel_i)\}_{\beta_i}$, $R' = R' \cup \{r^j: (v^j(\cdot), z^j)\}$}, $T = T - T_I - \beta_i \cdot T_R$ 
     \end{algorithmic}
\end{algorithmic}
\end{algorithm}
\end{figure}

\subsection{Querying for Rule Feedback}
\label{iws-method-rule-feedback}
After having extracted the set of $R_C$ candidate rules that cover $\segment_i$, we select up to $\beta$ candidate rules $r^j$ and query for their labels $z^j_i$, where $\beta$ is a hyper-parameter. 
Specifically, we first select in $R_C'$ all rules from $R_C$ that predict a label $z^j_i=y_i$ (thus agreeing with the expert's label for $s_i$). Then, we select from $R_C'$ the top $\beta$ rules with the highest precision (computed on $D_L$). 
Note that $R_C'$ might have fewer than $\beta$ rules in total, thus we use $\beta_i \leq \beta$ to indicate the number of rules selected finally.

Next, we query the labels $z^j_i$ for the $\beta_i$ selected rules at a cost of $\beta_i \cdot T_R$. 
At the end of the iteration, the $\beta_i$ labeled rules, which we denote as $\{(r^j, z^j)\}_{\beta_i}$, are added in $R$, where by design each rule $r^j$ will predict the same label $z^j=z^j_i$ for all instances that it covers. 
Our method ignores rules labeled with $z^j_i=\abstain$.

Throughout this interaction design, we assume that the domain expert can judge whether $r^j$ provides the correct label for most of the examples that the rule covers, and is aware that (i) a rule $r^j$ does not need to have perfect accuracy but rather represents a pattern that the expert intends to exploit to label examples more efficiently than manually; (ii) rule predictions will be aggregated to train a model in a noise-aware way. 
Similar to how expert-written rules are used for WSL, we assume that accepting a precise candidate rule for $\segment_i$ could improve the Student in the next iteration. This is possible, by augmenting the Student's training data with all the unlabeled examples covered by the rule, and by increasing the overlap of accepted rules $R$ on $D_U$, which provides useful signal for rule denoising, similar to inter-annotator agreement methods.

\subsection{Interactive Machine Teaching Algorithm}
\label{s:iws-interactive-machine-teaching-algorithm}
These steps outlined in Sections~\ref{iws-method-teacher-student-cotraining}-\ref{iws-method-rule-feedback} make up our interactive machine teaching method (Algorithm~\ref{algo:interactive-machine-teaching}), which we recap as follows.
First, our method clusters $D_U$ into hierarchical clusters.
In each interaction round: (1) we train the Teacher and Student using labeled data, unlabeled data, and expert-validated rules (line 3.1); (2) we apply the Student on unlabeled data to get soft labels (line 3.2); (3) we pick a candidate unlabeled instance (line 3.3) and obtain its instance label from an expert (line 3.4); (4) we extract candidate rules (line 3.5) and obtain the labels for $\beta_i$ rules from an expert (line 3.6); and (5) we update the labeled dataset, expert-validated rules, and the remaining budget (line 3.7).
In practice, we repeat Steps 3-6 (lines 3.3-3.6) in batches of 10 instances.
We repeat the full procedure until the budget $T$ runs out. 

By associating $r^j$ with a specific instance $s_i$, we give the expert extra context (e.g., the text of $\segment_i$) for deciding $z^j$.
Also, we hypothesize that, in practice, reading the text of the instance can help reduce the cost $T_R$ for deciding $z^j$. 
While some previous work assumes that labeling rules have no extra cost~\cite{poulis2017learning}, we assume that $T_R>0$.
The hyper-parameter $\beta_i$ controls how to distribute the budget $T$. Specifically, setting $\beta_i=0$ reduces to standard active learning, as \iwsysname will perform $\lfloor \frac{T}{T_I} \rfloor$ queries on instances only.
By setting $\beta_i \geq 1$, one can exploit feedback on rules that apply to $s_i$. 
As we will show, rule feedback leads to performance improvements relative to instance feedback only.

\begin{table*}[t]
\resizebox{2\columnwidth}{!}{
\begin{tabular}{lllllll}
\toprule
                  & \textbf{YouTube}             & \textbf{SMS}                                                            & \textbf{IMDB}                     & \textbf{Yelp}                                                                   & \textbf{TREC}                                                               & \textbf{AGNews}               \\\midrule
Classification task              & spam  & spam  & sentiment  & sentiment & question type & topic  \\
Domain            & user comments       & text messages                                                  & movies                   & reviews                                                                & web queries                                                        & news                 \\
\# Classes ($\numclasses$)  & 2                   & 2                                                              & 2                        & 2                                                                      & 6                                                                  & 4                    \\
Unlabeled size ($|D_U|$) & 1546 & 4531  & 19,960 & 30,360    & 4845  & 95,920 \\
Labeled train size ($|D_L|$) & 40 & 40  & 40 & 40    & 120   & 80 \\
Test size & 250                 & 500                                                            & 2500                     & 3800                                                                   & 500                                                                & 12,000                  \\
\# Prompt templates   & 5  & 5 & 15 & 12 & 6 & 9                    \\
\# Expert-provided rules ($R$)   & 10                  & 73                                                             & 5                        & 8                                                                      & 68                                                                 & 9                    \\
\bottomrule
\end{tabular}}
\caption{\small Statistics for available datasets with expert-labeled rules. }
\label{tab:iws-dataset-statistics}
\end{table*}

\section{Experimental Settings}
\label{s:iws-experimental-settings}
We now present our experimental setting for interactive machine teaching on several text classification datasets.

\paragraph{Datasets.} For our analysis and to evaluate our framework, we consider six benchmark datasets from diverse domains: (1) spam classification of YouTube comments~\cite{alberto2015tubespam}; (2) spam classification of SMS messages~\cite{almeida2011contributions}; (3) sentiment classification of IMDB movie reviews~\cite{maas2011learning}; (4) sentiment classification of Yelp reviews~\cite{zhang2015character}; (5) question classification from TREC-6~\cite{li2002learning}; and (6) topic classification in AGNews~\cite{zhang2015character}. 
Table~\ref{tab:iws-dataset-statistics} reports dataset statistics. 
For each dataset, we use expert-made rules that are provided by~\citet{zhang2021wrench} and prompt templates that are provided by~\citet{bach2022promptsource}.
For a fair comparison, we use exactly the same expert-written rules\footnote{All rules are described at \url{https://github.com/JieyuZ2/wrench}.} as in previous work, which can have various types such as keywords, regular expression patterns, and lexicons.

\paragraph{Experimental procedure.} 
To simulate the low-resource setting, we split the training examples into $D_L$ (labeled set) and $D_U$ (unlabeled set) by sampling 20 labeled examples per class ($20 \cdot \numclasses$ in total) uniformly at random, which we use in $D_L$, while we use the rest in $D_U$. 
To be consistent with our low-resource assumptions, we downsample the validation set  (used for training Student via early stopping) to match the size of $D_L$.
For interactive approaches, we consider the extreme low-resource setting where $R=\emptyset$.
We simulate expert feedback for candidate instances $s_i$ from $D_U$ (Section~\ref{iws-method-instance-feedback}) using the ground-truth labels of $D_U$ (hidden to the main algorithm), which is common in active learning research~\cite{zhang-etal-2022-survey}.
We simulate expert feedback for candidate automatic rules (Section~\ref{iws-method-rule-feedback}) using all ground-truth labels in $D_U$: a candidate rule $r^j$ is accepted if it correctly classifies more than $t_{oracle}$ of the instances in $D_U$ that it covers. 
We experiment with different values of $t_{oracle}$: $25\%$, $50\%$, $75\%$, $90\%$, and $100\%$ and study their impact on the student's accuracy.

For a robust evaluation, for each method we run 10 different experiments with different random seeds, thus each run corresponds to a different version of $D_L$, $D_U$, and $R$. 
We report the average test performance over the 10 different runs. 
As evaluation metric, we use the macro-averaged F1 of the student model on the test set.

\paragraph{Model configuration.} 
For a fair comparison, we use exactly the same text pre-processing (tokenization, embedding) as in the WRENCH benchmark~\cite{zhang2021wrench}. 
Following~\citet{zhang2021wrench}, we represent each text instance ($\segment_i$) as a vector using pre-trained BERT~\cite{devlin2019bert}, specifically as the output embedding of the [CLS] token of BERT-base\footnote{\url{https://huggingface.co/google-bert/bert-base-cased}}.
For the hyper-parameters and search space for bag-of-words logistic regression, multilayer perceptron, and BERT, see Table 10 in~\citet{zhang-etal-2022-survey}.
For candidate rule extraction, we consider conjunctions (AND) of up to $t_{len}=3$ features consisting of $n$-grams with $n=1, 2, 3$; linguistic features (part-of-speech tags and named entities extracted using the spaCy library\footnote{\url{https://spacy.io/usage/linguistic-features/}}); and prompt-based features as the top $k=10$ tokens predicted by pre-trained RoBERTa~\citep{liu2019roberta} for each of the templates provided by~\citet{bach2022promptsource}\footnote{Prompt templates are available at \url{https://github.com/bigscience-workshop/promptsource.}}.
For our analysis of rule characteristics, we experiment with different values for the minimum rule coverage on $D_U$ ($t_{cov} \in \{10, 100, 1000\}$) and the minimum rule precision based on $D_L$ ($t_{prec} \in \{25\%, 50\%, 75\%, 100\%\}$).
In \iwsysname, we use $t_{cov}=100$ and $t_{prec}=75\%$.
For interaction, we study different relative values for $\beta$ (maximum number of rules per instance), $T_R$ (rule labeling cost) and $T_I$ (instance labeling cost). 

\paragraph{Model comparison.} 
For a robust evaluation of our approach, we compare several approaches that utilize different resources:
\begin{enumerate}
    \item \textbf{``Fully supervised'':} a model trained in the high-resource setting using \emph{all} labeled data. 
    \item \textbf{``Low supervised'':} a model trained in the low-resource setting using only $D_L$. 
    \item \textbf{``Semi supervised'':} a model trained using $D_L$ and $D_U$. We consider self-training~\cite{nigam2000analyzing,lee2013pseudo} for up to 25 iterations with early stopping based on the validation performance. 
    \item \textbf{``WSL'':} a model trained using $D_L$, $D_U$, and $R$. We experiment with different methods, including unweighted majority voting and weighted aggregation of rule predictions with majority voting, Snorkel~\cite{ratner2017snorkel},
    Dawid-Skene~\cite{dawid1979maximum}, FlyingSquid~\cite{fu2020fast}, MeTaL~\cite{ratner2019training}, and ASTRA~\cite{karamanolakis2021astra}.
    \item \textbf{``Active learning'':} a model trained using $D_L$, $D_U$, and the interaction budget $T$. We experiment with standard active learning (performing $\lfloor \frac{T}{T_I} \rfloor$ queries on instances only) with different acquisition functions, including random instance selection, uncertainty-based sampling, hierarchical sampling~\cite{dasgupta2008hierarchical}, and contrastive active learning~\cite{margatina2021active}. We also evaluate IWS~\cite{boecking2020interactive}, which considers $n$-gram rule families and performs $\lfloor \frac{T}{T_R} \rfloor$ queries on rules only.\footnote{Unfortunately, the code repository for PRBoost~\cite{zhang2022prompt}, \url{https://github.com/rz-zhang/PRBoost}, does not contain any code as of August 9th, 2024.} 
    \item \textbf{``\iwsysname''}: a model trained using our interactive machine teaching method that uses $D_L$ and $D_U$, and spends the interaction budget $T$ to perform queries on both instances and rules. 
\end{enumerate}

For a fair comparison, we use exactly the same modeling configuration across all methods (see pagraph ``Model configuration'' for details). 

\begin{figure*}[t]
    \centering
    \begin{subfigure}[t]{0.49\textwidth}
        \centering
        \includegraphics[width=\columnwidth]{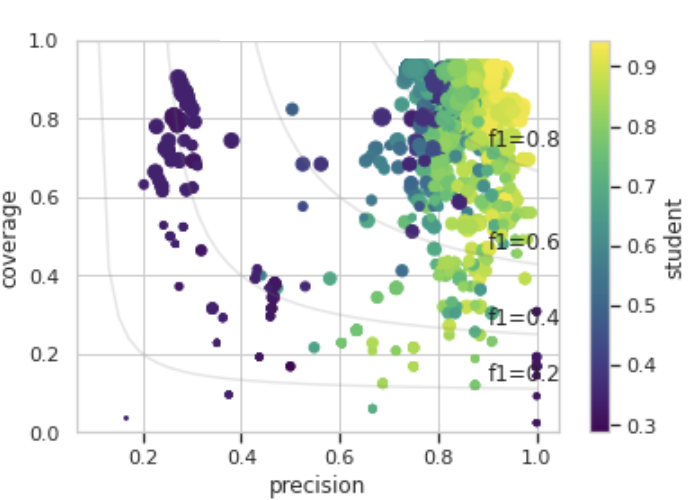}
        \caption{YouTube.}
        \label{fig:youtube_offline_scatterplot}
    \end{subfigure}%
    \begin{subfigure}[t]{0.49\textwidth}
        \centering
                \includegraphics[width=\columnwidth]{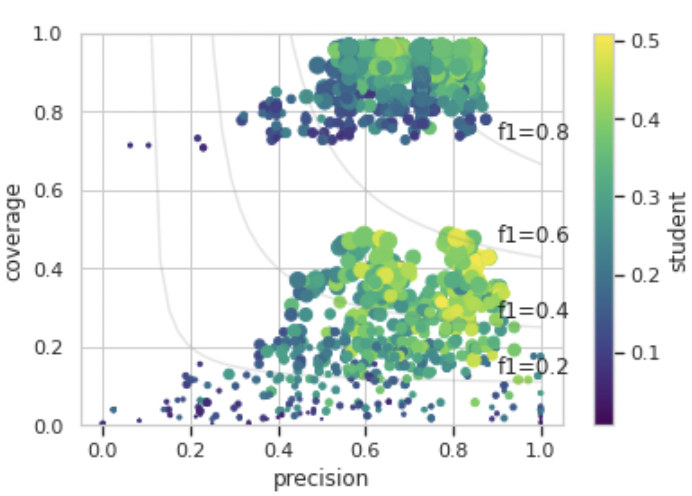}
        \caption{TREC.}
        \label{fig:trec_offline_scatterplot}
    \end{subfigure}
    \caption{Precision-coverage scatterplots reporting the precision (x-axis) and coverage (y-axis) of the Teacher. Each data point corresponds to a different Teacher-Student pair and its color indicates the F1 score of the Student.}
    \label{fig:teacher_student_offline_scatterplots}
\end{figure*}

\begin{table*}[t]
    \centering
    \begin{tabular}{l|c|c|c|c|c|c}
    \toprule
         &  \textbf{YouTube} & \textbf{SMS} & \textbf{Yelp} & \textbf{IMDB} & \textbf{TREC} & \textbf{AGNews}\\
    \midrule
    Coverage weight &  0.20 & 0.00 & 0.22 & 0.23 & 0.30 & 0.46 \\
    Precision weight & \textbf{0.80} & \textbf{1.00} & \textbf{0.78} & \textbf{0.77} & \textbf{0.70} & \textbf{0.54} \\    
    \bottomrule
    \end{tabular}
    \caption{Quantifying the relative importance of Teacher coverage and precision for training an accurate Student. Across all datasets, precision has a higher weight than coverage.}
    \label{tab:iws_precision_vs_coverage_weights}
\end{table*}
\section{Experimental Results}
\label{s:iws-experimental-results}
We now present our analysis of expert-provided rules (Section~\ref{s:analysis-human-labeled-rules}), results on automatic rule extraction (Section~\ref{s:experiments-automatic-rule-suggestion}), and our experiments for interactive machine teaching with queries on instances and rules (Section~\ref{s:experiments-interactive-machine-teaching}).

\subsection{Analysis of Expert Rules}
\label{s:analysis-human-labeled-rules}
In this section, we analyze existing datasets with expert-labeled rules and simulate low-resource rule settings to understand the impact of Teacher properties on the performance of the Student. 

\paragraph{Analysis of the precision vs.~coverage trade-off.}
In Section~\ref{s:motivation-interactive-weak-supervision}, we highlighted one challenging question: should one prioritize rules that cover more examples but have a relatively lower precision or a few rules that have higher precision but lower coverage?
To analyze the precision-coverage trade-off, we create different Teacher versions using different subsets of the expert-labeled rules and evaluate the performance of Student using each Teacher separately.
For a robust analysis, we evaluate multiple Teacher types (majority voting, Snorkel~\cite{ratner2016data}, Dawid-Skene~\cite{dawid1979maximum}, MeTaL~\cite{ratner2019training}, FlyingSquid~\cite{ratner2019training}), and multiple Student types (bag-of-words logistic regression, multilayer perceptron, BERT). See Section~\ref{s:iws-experimental-settings} for implementation details.
For each Teacher type, we keep different randomly-selected subsets of the rules in $R$ ranging from 1\% to 100\%. 
For each Teacher-Student combination, we run 10 different experiments with different random seeds. 
This results in more than 1,000 Teacher-Student configurations for each dataset. 

\begin{table*}[t]
\centering 
\resizebox{1.8\columnwidth}{!}{
\begin{tabular}{llllllllllll}
\toprule \textbf{Rule family} & \textbf{YouTube} & \textbf{SMS}  & \textbf{IMDB} & \textbf{Yelp} & \textbf{TREC} & \textbf{AGNews} & \textbf{AVG F1}  \\
\midrule
Expert &   \textbf{90.0}   & 86.8 & 71.2 & 80.2 & \textbf{57.0} & 75.9  & 76.8 \\
Automatic ($n$-gram;~\citet{boecking2020interactive})  & 76.4 &	79.7&	49.1 &	54.9	& 52.7	& 74.8	& 64.6\\
Automatic (ours) & 82.7 &	\textbf{91.4}	& \textbf{73.5} &	\textbf{86.1} &	53.3	& \textbf{78.1}	& \textbf{77.5} \\
\bottomrule
\end{tabular}}
\caption{\small F1 score of the WSL method trained with expert rules and automatically extracted rules from two different families, namely, $n$-gram rules and high-level rules (conjunctions of $n$-grams, named entities, and prompt-based features). Our automatic rules lead to better performance than expert rules and $n$-gram rules. Performance differences for each dataset are statistically significant at $p < 0.05$ using the Student's t-test.}
\label{s:iws-rule-extraction-results}
\end{table*}

\begin{figure}
    \centering
    \includegraphics[width=\columnwidth]{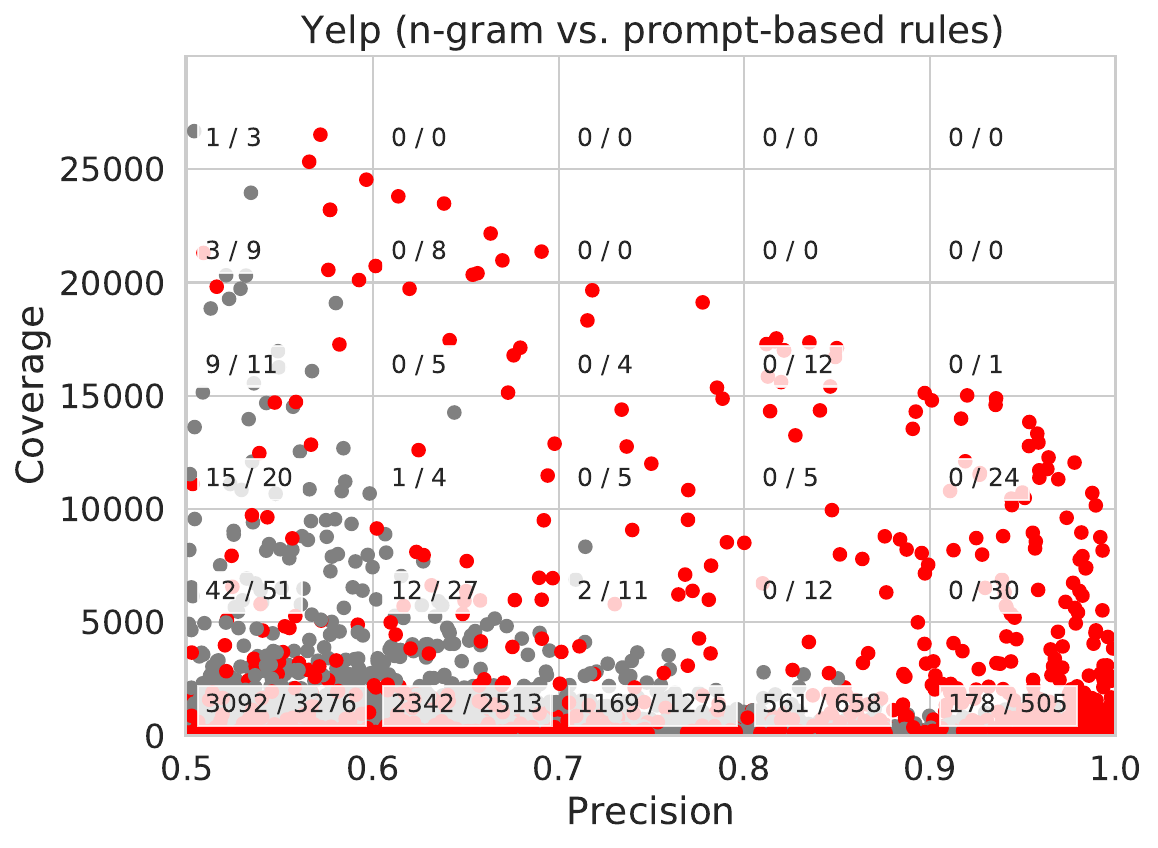}
    \caption{\small Precision-coverage scatterplots for automatically extracted $n$-grams (grey) and prompt-based rules (red). Grid numbers show the count of $n$-gram/prompt-based rules on the corresponding grid. Prompt-based rules can achieve relatively higher precision and coverage than $n$-gram rules.}
    \label{fig:autorule_scatterplots}
\end{figure}
Figure~\ref{fig:teacher_student_offline_scatterplots} summarizes the results across all experiments for YouTube and TREC.
While different datasets have Teacher-Student pairs with different characteristics, there are patterns that are prevalent across datasets. 
First, a more accurate Teacher does not necessarily lead to a more accurate Student. 
For example, in YouTube (Figure~\ref{fig:teacher_student_offline_scatterplots}) some Teachers with F1 $\geq 0.6$ train a Student with F1 $\geq 0.5$, while other Teachers with F1 $\leq 0.2$ train a Student with F1 $\geq 0.8$.
This result implies that naively optimizing the Teacher's performance (according to the standard ``data programming'' paradigm~\cite{ratner2016data}) might not lead to the best performing student model.

A second pattern that is prevalent across datasets is that the Teacher's precision is more important than coverage for training an accurate Student. 
In the scatterplots of Figure~\ref{fig:teacher_student_offline_scatterplots}, most Teachers with high precision train high-quality Students, while many Teachers with high coverage train low-quality Students. 
To quantify this observation, we compute precision-coverage weights using the Teacher's precision and coverage to predict the Student's F1 score. 
Specifically, we compute the Student's F1 score as the weighted geometric average of the Teacher's precision and coverage, and we tune the corresponding weights using grid search. 
A higher weight thus indicates that the corresponding feature is more important for the prediction of the Student's F1 score.
Table~\ref{tab:iws_precision_vs_coverage_weights} shows the estimated precision and coverage weights for all datasets.
Across all datasets, precision has higher weight than coverage: more precise Teachers lead to more accurate Students. 

Our observation that rule precision is more important than coverage explains recent design choices for WSL~\cite{awasthi2019learning,hsieh2022nemo}, such as the ``contextualized LF modeling'' component of~\citet{hsieh2022nemo}, which explicitly reduces rule coverage to improve rule precision.
Moreover, our observation might inform guidelines for rule creation. 
In YouTube, for instance, if we reject all Teacher models with coverage lower than 0.5, then the precision's importance weight increases from 0.75 to 0.84, indicating that focusing on precision would be beneficial. 
Therefore, one potential guideline is that if the Teacher has a coverage higher than 50\%, then the main focus should be on improving its precision. 

\begin{table*}[t]
\resizebox{2\columnwidth}{!}{
\begin{tabular}{lllllccccccc}
\toprule
\textbf{Method} & $|D_L|$   & $D_U$ & $R$    & $T$ ($T_I$, $T_R$)      & \textbf{YouTube} & \textbf{SMS}  & \textbf{IMDB} & \textbf{Yelp} & \textbf{TREC} & \textbf{AGNews} & \textbf{AVG F1}  \\
\midrule
Fully Supervised & 100\%  & -    & -     & -           & 94.0    & 95.6 & 79.6 & 87.5 & 90.3 & 80.7   & 88.0 \\\hline 
Low  Supervised  & 20$\cdot\numclasses$ & -    & -     & -           & 79.8    & 82.5 & 61.6 & 70.4 & 55.0 & 58.8   & 68.0 \\
Semi Supervised  & 20$\cdot\numclasses$ & \checkmark  & -     & -           & 80.7    & 83.2 & 63.4 & 72.0 & 55.0 & 60.7   & 69.2 \\

WSL (ASTRA) & 20$\cdot\numclasses$ & \checkmark  & \checkmark& -           & \textbf{90.0}    & 86.8 & \textbf{71.2} & 80.2 & 57.0 & \textbf{75.9}   & \textbf{76.8} \\
Active Learning (hierarchical) & 20$\cdot\numclasses$ & \checkmark  & -     & 100 (100, 0)    & 85.3    & \textbf{89.9} & 67.6 & \textbf{81.2} & \textbf{61.4} & 71.4   & 76.1\\

\iwsysname & 20$\cdot\numclasses$ & \checkmark  & -  & 100 (50, 50) &  \textcolor{red}{\textbf{91.4}} &  \textcolor{red}{\textbf{94.8}} &	 \textcolor{red}{\textbf{79.3}} &	 \textcolor{red}{\textbf{86.2}} &	 \textcolor{red}{\textbf{66.6}} &	 \textcolor{red}{\textbf{78.8}} &	 \textcolor{red}{\textbf{82.8}}\\

\bottomrule
\end{tabular}}
\caption{\small F1 score reported for various methods on 6 datasets. Columns 2-5 describe the resource usage, specifically the size of labeled set $D_L$ (where the number of classes $K$ varies per dataset), the usage of the unlabeled set $D_U$ and initial expert rules $R$, and the interaction budget for feedback on instances ($T_I$) and rules ($T_R$). We report the best performing method for each category. \iwsysname outperforms WSL and Active Learning across all datasets, where performance differences for each dataset are statistically significant at $p < 0.05$ using the Student's t-test.}
\label{s:iws-main-results-table}
\end{table*}

\subsection{Analysis of Automatic Rules}
\label{s:experiments-automatic-rule-suggestion}
In this section, we compare our rule family to $n$-gram rules and expert rules. 
Figure~\ref{fig:autorule_scatterplots} shows precision-coverage scatterplots for rules automatically extracted by our method.
For this analysis, we have included all rules with precision higher than 0.5 and coverage higher than 0. 
Rules with high-level predicates (conjunctions of $n$-grams, named entities, and prompt-based features) can achieve relatively high precision and coverage compared to $n$-gram predicates and thus are promising to improve the overall performance of interactive machine teaching.

Table~\ref{s:iws-rule-extraction-results} reports the performance of ``WSL'' with automatically extracted rules extracted by our method using $t_{cov}=100$ (minimum coverage) and $t_{prec}=0.75$ (minimum precision). Across all datasets, our rule family is more effective than $n$-gram rules and could thus improve the effectiveness of automatic rule extraction.
Also, across most datasets (except TREC and YouTube), our rule family is more effective than expert-provided rules: we effectively use $D_U$ and $D_L$ to discover high-quality rules. 
As an exception, TREC contains the highest number of manually-crafted rules compared to the rest of the datasets. 
As we will show next, expert interaction can lead to further improvements.

\subsection{Interactive Machine Teaching}
\label{s:experiments-interactive-machine-teaching}
Table~\ref{s:iws-main-results-table} reports classification results of different methods for each dataset. 
For brevity, we report the best method under each category and list the average F1 across datasets (see AVG F1 column). In interactive methods, we assume $T_R=T_I$ and fix $\beta=1$ (while we study different values later).

\paragraph{Non-interactive approaches.} Across non-interactive approaches, WSL ASTRA performs best: using both labeled instances and expert-provided rules is more effective than using just labeled instances (in Low Supervised or Semi Supervised), which agrees with conclusions from recent work~\cite{karamanolakis2021astra}.
ASTRA outperformed other WSL methods, including majority voting (AVG F1$=74.1$) and Snorkel (AVG F1$=74.5$).

\paragraph{Active learning approaches.}
Using the extra interaction budget $T$ in Active Learning improves over Low Supervised: labeling extra instances leads to important performance boosts, as expected.
Hierarchical sampling performs better than random sampling (AVG F1 = 75.0), uncertainty-based sampling (AVG F1 = 75.3), contrastive active learning (AVG F1 = 74.1), and IWS (AVG F1 = 75.3).
For SMS, Yelp, and TREC, Active Learning with a budget of $T=100$ outperforms ASTRA: acquiring 100 extra instance labels is more effective than collecting expert rules for these datasets.
However, for YouTube, IMDB, and AGNews, Active Learning (hierarchical) does not outperform ASTRA, which highlights that expert-provided rules are worth many examples.
The above results suggest that there is no clear winner between Active Learning and WSL, and their relative performance varies across datasets.

\paragraph{Interactive learning with queries on rules and instances.}
In Table~\ref{s:iws-main-results-table}, \iwsysname with a budget of $T=100$ performs better than the best Active Learning (hierarchical) approach with the same budget: leveraging feedback on both instances and rules within a limited budget is more effective than feedback on instances only. 
Interestingly, even without using any expert-provided rules, \iwsysname outperforms ASTRA.
This indicates that automatically-generated rules (analyzed in Section~\ref{s:experiments-automatic-rule-suggestion}) are effective.
While the ASTRA Student might capture implicit rules via self-training, many rules could be inaccurate, thus highlighting the importance of expert interaction.

Table~\ref{s:iws-all-comparisons} summarizes the results for all methods and ablation experiments. \iwsysname performs better than its ablations without instance labeling (by 6\%) and without rule labeling (by 8\%): feedback on both instances and rules is the most effective. 
Also, our rule family is more effective than its ablations without $n$-gram rules (by 4\%) and without prompt-based rules (by 3\%).
Performance differences on each dataset are statistically significant at $p < 0.05$ using the Student's t-test.

\begin{figure}
    \centering
    \begin{subfigure}[b]{\columnwidth}
    \includegraphics[width=\columnwidth]{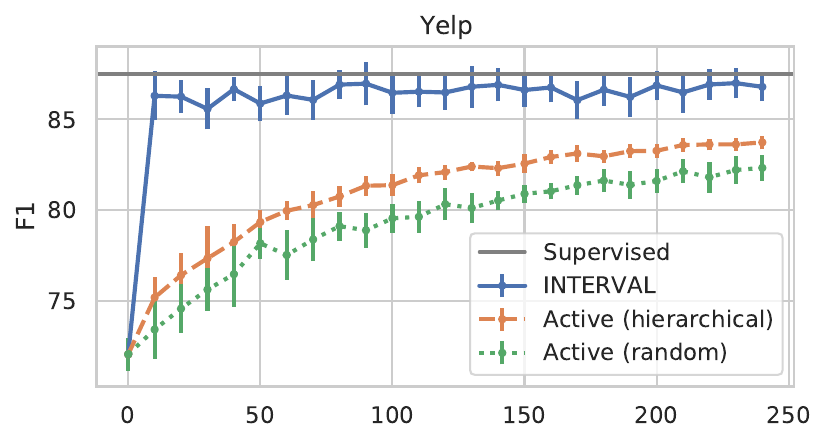}
    \end{subfigure}
    \begin{subfigure}[b]{\columnwidth}
    \includegraphics[width=\columnwidth]{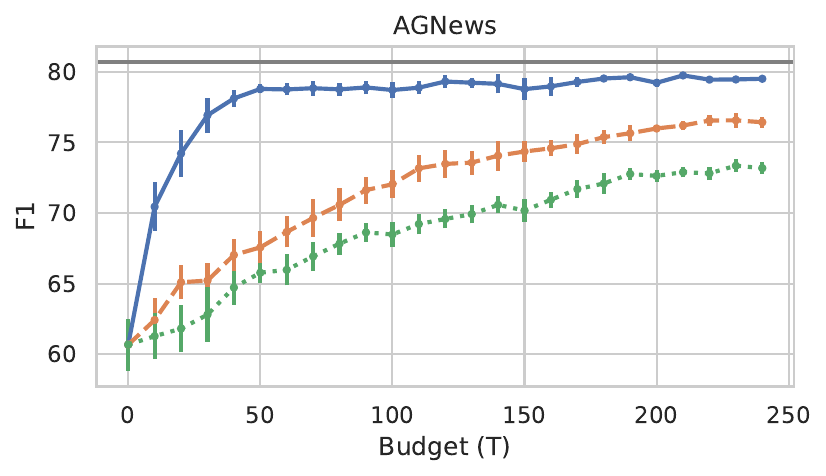}
    \end{subfigure}
    \caption{\small Performance of interactive methods on Yelp (top) and AGNews (bottom) as a function of budget ($T$). \iwsysname outperforms Active Learning with strongest improvements in low-budget settings (left).}
    \label{fig:interval_yelp_plot}
\end{figure}

\paragraph{Performance with different budget values.}
Table~\ref{s:active-learning-varying-budget} reports the performance of interactive methods with different budget sizes ranging from 10 to 250. 
\iwsysname requires as few as $T$=10 queries to reach F1 values that existing active learning methods cannot match even with $T$=100 queries.
Figure~\ref{fig:interval_yelp_plot} shows the performance of \iwsysname compared to Active Learning approaches on Yelp and AGNews. 
\iwsysname leads to a big performance boost especially in low-budget settings where $T<100$.
Our results highlight that \iwsysname can effectively leverage feedback on both instances and automatic rules, and outperform previous interactive methods.

\begin{table}[t]
\resizebox{\columnwidth}{!}{
\begin{tabular}{ll}
\toprule
\textbf{Method} &  \textbf{AVG F1}  \\
\midrule
Fully Supervised &88.0 \\\hline 
Low  Supervised & 68.0 \\
Semi Supervised (self-training)~\cite{lee2013pseudo}  & 69.2 \\
WSL  (majority voting) &74.0 \\
WSL  (Snorkel)~\cite{ratner2017snorkel} &  74.2 \\
WSL  (FlyingSquid)~\cite{fu2020fast} &  74.2 \\
WSL  (MeTaL)~\cite{ratner2019training} &  74.7 \\
WSL (ASTRA)~\cite{karamanolakis2021astra} &  76.8 \\
Active Learning (random)& 75.0 \\
Active Learning (uncertainty) &  75.3 \\
Active Learning (contrastive)~\cite{margatina2021active}&  75.4 \\
Active Learning (hierarchical)~\cite{dasgupta2008hierarchical} & 76.1\\
Interactive Rule Labeling (IWS)~\cite{boecking2020interactive}&  75.1 \\
\iwsysname &  \textcolor{black}{\textbf{82.8}}\\\midrule
\iwsysname w/o instance labeling & 78.2 $\downarrow$6\% \\
\iwsysname w/o rule labeling & 76.1 $\downarrow$8\% \\
\iwsysname w/o prompt-based rules& 79.7 $\downarrow$4\% \\
\iwsysname w/o n-gram rules& 80.2 $\downarrow$3\%\\
\bottomrule
\end{tabular}}
\caption{\small Comparison of all methods (average F1 across datasets) and ablation experiments.}
\label{s:iws-all-comparisons}
\end{table}

\begin{table}[t]
\resizebox{\columnwidth}{!}{
\begin{tabular}{lcccccc}
\toprule
& \multicolumn{6}{c}{\textbf{Budget ($T$)}}\\

\textbf{Method} & 10 & 50 & 100 & 150 & 200 & 250 \\
\midrule
Active Learning (rand.)& 68.1 & 71.8 & 75.0 & 76.5& 78.0& 78.4\\
Active Learning (hier.) & 68.4& 73.9 & 76.1 & 78.3& 79.3& 79.9\\
\iwsysname & \textbf{76.2} &\textbf{81.1} & \textbf{82.8} & \textbf{84.3} & \textbf{85.5}& \textbf{86.2}\\
\bottomrule
\end{tabular}}
\caption{\small Comparison of interactive methods (average F1) with different budget sizes ($T$). Performance differences for each budget size are statistically significant at $p < 0.05$ using the Student's t-test.}
\label{s:active-learning-varying-budget}
\end{table}

\paragraph{Evaluating the relative cost of rules and instances.}
So far, we have evaluated our method by assuming that $T_R=T_I$. Here, we experiment with different relative costs of labeling rules ($T_R$) and instances ($T_I$). We assume $T = 100 \cdot T_I$ (fixed total budget), $\beta=1$ (labeling up to one rule per instance), and find the maximum value for $T_R$ so \iwsysname ($T=\sum_i T_I + \beta_i \cdot T_R$) has an F1 score that is at least as high as the best Active Learning (hierarchical) method ($T=\sum_i T_I$).
Table~\ref{tab:relative_cost_analysis} reports the maximum $T_R$ value for each dataset. On average across datasets, feedback on rules and instances is more effective than feedback on instances as long as $T_R \leq 5.2 T_I$, though this value varies significantly per dataset and can be as high as $9T_I$ (for Yelp).
In other words, our hybrid method for labeling rules and instances is highly effective even when labeling rules is 9 times (for Yelp) more expensive than labeling instances.

\paragraph{How many rules to label per instance.}
Table~\ref{s:rules-per-instance} shows the performance of $\iwsysname$ by varying $\beta$ (maximum number of rules to label per instance). Labeling up to one rule ($\beta=1$) gives strong boosts compared to no rule labeling ($\beta=0$) across datasets while labeling up to two rules ($\beta=2$) gives further improvements in some tasks (YouTube, Yelp, AGNews).
However, increasing $\beta$ to values higher than 2 is less effective: when $\beta$=5, then either less-accurate or redundant rules are queried, while this interaction budget could be used more effectively by labeling more instances (and the associated rules).
Table~\ref{tab:example-interval-trec} shows an example from AGNews (classes are ``World,'' ``Sports,'' ``Business,'' and ``Sci/Tech'') where $\iwsysname$ is applied with $\beta=5$.
The candidate instance is labeled as ``World'' topic and out of the $\beta_i=3$ rules that were queried (by satisfying the minimum precision and coverage thresholds), 2 were accepted and 1 was rejected as ``international'' also appears in other topics (e.g., ``Business'').  
Our analysis suggests that most performance benefits are realized by labeling up to 1 rule per instance, while future research could dynamically determine the threshold $\beta$, for example as a function of task characteristics and labeling costs.

\begin{table}[t]
    \centering
    \resizebox{\columnwidth}{!}{
    \begin{tabular}{lcccccc} 
    \toprule
         & \textbf{YouTube} & \textbf{SMS} & \textbf{IMDB} & \textbf{Yelp} & \textbf{TREC} & \textbf{AGNews} \\
       $T_R$   & 4 $T_I$ & 4 $T_I$ & 8 $T_I$ & 9 $T_I$& 2$T_I$ & 4 $T_I$\\
      \bottomrule
    \end{tabular}}
    \caption{\small Maximum value for $T_R$ (cost of rule feedback) as a function of $T_I$ (cost of instance feedback) so that feedback on rules and instances is more effective than just instance feedback.}
    \label{tab:relative_cost_analysis}
\end{table}

\begin{table}[t]
\resizebox{\columnwidth}{!}{
\begin{tabular}{clllllll}
\toprule
\textbf{$\beta$} & \textbf{YouTube} & \textbf{SMS}  & \textbf{IMDB} & \textbf{Yelp} & \textbf{TREC} & \textbf{AGNews} & \textbf{AVG F1}  \\
\midrule
0& 85.3    & 89.9 & 67.6 & 81.2 & 61.4 & 71.4   & 76.1\\
1 &91.4 &  \textbf{94.8} &	 \textbf{79.3} &	 86.2 &	 \textbf{66.6} &	 78.8 &	 82.8* \\
2&  \textbf{91.9}&	\textbf{94.8}&	79.2*&	\textbf{87.3}&	65.0&	\textbf{79.7}& \textbf{83.0}\\
5 &  91.0 &	94.7*&	78.4&	86.9	&62.5	&79.2& 82.1\\
\bottomrule
\end{tabular}}
\caption{\small F1 score of $\iwsysname$ for each dataset by varying $\beta$ (maximum rules labeled per instance). An asterisk (*) next to a number denotes that the difference is not statistically significant when compared to the bolded values, as determined by a p-value greater than 0.05 using the Student’s t-test.}
\label{s:rules-per-instance}
\end{table}

\section{Discussion and Future Work}
\label{s:interval_discussion}
Our framework and analysis demonstrates the advantages of soliciting feedback on both candidate rules and individual instances. We identify several areas for future research and discuss them next.

As future work, we will explore additional design choices for \iwsysname, including instance selection strategies (e.g., based on rule informativeness), rule extraction methods (e.g., based on rule diversity), and weak supervision techniques.  
While \iwsysname selects up to $\beta$ candidate rules per instance (where $\beta_i$ depends on how many rules satisfy the precision and coverage thresholds), we could further explore
adaptive querying protocols, for example dynamically determining $\beta$ or selectively skipping instance labeling based on dataset characteristics or labeling costs.
We could also extend \iwsysname to support richer types of feedback, such as editing (rather than accepting or rejecting) candidate rules and prompt templates (rather than relying on fixed templates from~\citet{bach2022promptsource}).
More research is required from a user perspective, for example on how to visualize rules~\cite{lertvittayakumjorn-etal-2022-grasp} and effectively present a combination of rules and instances for expert labeling.
\iwsysname supports prompting pre-trained models just for training data creation, and can work with any model for inference, thus enabling applications where deploying large language models might not be possible. 
We expect further gains by creating rules using more powerful pre-trained models such as  InstructGPT~\citep{ouyangtraining}; PaLM-T5~\citep{chung2022scaling}; LLaMA~\citep{touvron2023llama,touvron2023llama2}. 
We also expect performance improvements by replacing the Student using stronger pre-trained models and by representing instances using more recent text embedding techniques~\citep{he2020deberta,wang2023improving,su-etal-2023-one,muennighoff2024generative}.
INTERVAL could also be extended for multi-label classification by changing the Teacher-Student co-training objective (Section~\ref{iws-method-teacher-student-cotraining}) and for other broader tasks by generating rules from more complex rule families using models such as Toolformer~\citep{schick2023toolformer}.

\begin{table}[t]
    \centering
    \resizebox{\columnwidth}{!}{
    \begin{tabular}{l}
    \toprule
     \textbf{Text instance $\segment_i$}: \\
    ``\textit{Prime Minister Manmohan Singh today said international} \\
     \textit{environment for India's development was highly favourable...}''\\
    \textbf{Queries}: \\
    - Instance label: World\\
    - Rule 1: NGRAM=``prime minister'' $\rightarrow$ World \textcolor{forestgreen}{\checkmark}\\
    - Rule 2: PROMPT\_IS\_ABOUT=``politics'' $\rightarrow$ World \textcolor{forestgreen}{\checkmark}\\
    - Rule 3: NGRAM=``international'' $\rightarrow$ World \textcolor{red}{\ding{55}}  \\
    - Rule 4: - \\
    - Rule 5: -\\
     \bottomrule
    \end{tabular}}
    \caption{\small Example from AGNews with $\beta=5$.  All classes are ``World,'' ``Sports,'' ``Business,'' and ``Sci/Tech.'' Out of the rules that were queried, 2 were accepted and 1 was rejected.} 
    \label{tab:example-interval-trec}
\end{table}

Our current experimental evaluation used simulated expert feedback, because a definitive evaluation involving actual subject matter experts would be too expensive. 
A potential stopgap is to use large language models (such as ChatGPT), which may be too expensive to query at test time, but are cheaper than subject matter experts to query at training time for selected instances.

\section{Conclusions}
\label{iws:conclusions}
In this paper, we presented an interactive machine teaching approach that queries experts for feedback on both instances and automatically generated rules. 
Our findings show that, even though rules are domain specific and have diverse characteristics, there are patterns that are prevalent across datasets.
Specifically, a higher-F1 Teacher does not necessarily lead to a higher-F1 Student. We identified that the Teacher's precision is more important than coverage for training an accurate Student.
These findings could potentially inform guidelines for rule creation. 
Our analysis demonstrates that automatic rules based on high-level predicates are more accurate than rules based on $n$-gram predicates. 
We additionally showed that by asking queries on both instances and automatically extracted rules, our method can be more effective than active learning methods. 

\newpage

\section*{Acknowledgments}
We thank the reviewers and action editors for their constructive feedback. This material is based upon work supported by the National Science Foundation under Grant No. IIS-15-63785.
\bibliography{references}
\bibliographystyle{acl_natbib}

\end{document}